\documentclass[10pt,twocolumn,letterpaper]{article}

\usepackage{cvpr}              % To produce the CAMERA-READY version
\usepackage{subcaption} 
\usepackage{colortbl}
\definecolor{verylightgray}{gray}{0.9}
\definecolor{beaublue}{rgb}{0.74, 0.83, 0.9}

\newcommand{\method}{DeCafNet\xspace}
\newcommand{\grounding}{DeCaf-Grounder\xspace}

\newcommand{\headline}[1]{\smallskip \noindent \textbf{#1}}

\newcommand{\f}{\mathbf{f}}
\newcommand{\q}{\mathbf{q}}
\newcommand{\cls}{\mathbf{q}_\text{cls}}

\newcommand{\F}{\mathbf{F}}
\newcommand{\Z}{\mathbf{Z}}
\newcommand{\Zl}{\mathbf{Z}^l}
\newcommand{\bS}{\mathbf{S}}
\newcommand{\bpl}{\mathbf{p}^l}
\newcommand{\bhpl}{\hat{\mathbf{p}}^l}
\newcommand{\bhP}{\hat{\mathbf{P}}}
\newcommand{\bH}{\mathbf{H}}
\newcommand{\bUl}{\mathbf{U}^l}

\definecolor{cvprblue}{rgb}{0.21,0.49,0.74}
\usepackage[pagebackref,breaklinks,colorlinks,allcolors=cvprblue]{hyperref}
\usepackage{soul}
\usepackage[accsupp]{axessibility}
\def\paperID{10657} % *** Enter the Paper ID here
\def\confName{CVPR}
\def\confYear{2025}

\title{\method: Delegate and Conquer for Efficient Temporal Grounding\\in Long Videos}

\author{Zijia Lu\textsuperscript{1,2}$\dagger$\thanks{This work was conducted during Z. Lu’s internship at Microsoft. Z. Lu and E. Elhamifar were supported in part by DARPA PTG (HR00112220001), NSF (IIS-2115110) and ARPA-H (1AY2AX000062) during the preparation of the manuscript. Content does not necessarily reflect the position/policy of the Government.}, 
\hspace{5pt} A S M Iftekhar\textsuperscript{1}$\dagger$, 
\hspace{5pt} Gaurav Mittal\textsuperscript{1}, 
\hspace{5pt} Tianjian Meng\textsuperscript{1}, 
\hspace{5pt} Xiawei Wang\textsuperscript{1}, \\ 
\hspace{5pt} Cheng Zhao\textsuperscript{1}, 
\hspace{5pt} Rohith Kukkala\textsuperscript{1}, 
\hspace{5pt} Ehsan Elhamifar\textsuperscript{2}, 
\hspace{5pt} Mei Chen\textsuperscript{1} \\ 
Microsoft\textsuperscript{1}, \hspace{5pt} Northeastern University\textsuperscript{2} \\
{\tt\small lu.zij@northeastern.edu, $\;$ \{asmiftekhar, gamit, tianjianmeng, xiaweiwang\}@microsoft.com} \\
{\tt\small \{chengzhao, rokukkal, meic\}@microsoft.com, e.elhamifar@northeastern.edu}
}

\begin{document}
\maketitle
\def\thefootnote{$\dagger$}\footnotetext{Equal contribution}\def\thefootnote{\arabic{footnote}}
\begin{abstract}

Long Video Temporal Grounding (LVTG) aims at identifying specific moments within lengthy videos based on user-provided text queries for effective content retrieval. The approach taken by existing methods of dividing video into clips and processing each clip via a full-scale expert encoder is challenging to scale due to prohibitive computational costs of processing a large number of clips in long videos. To address this issue, we introduce DeCafNet, an approach employing ``delegate-and-conquer'' strategy to achieve computation efficiency without sacrificing grounding performance. DeCafNet introduces a sidekick encoder that performs dense feature extraction over all video clips in a resource-efficient manner, while generating a saliency map to identify the most relevant clips for full processing by the expert encoder. To effectively leverage features from sidekick and expert encoders that exist at different temporal resolutions, we introduce DeCaf-Grounder, which unifies and refines them via query-aware temporal aggregation and multi-scale temporal refinement for accurate grounding. Experiments on two LTVG benchmark datasets demonstrate that DeCafNet reduces computation by up to 47\% while still outperforming existing methods, establishing a new state-of-the-art for LTVG in terms of both efficiency and performance. 
Our code is available at \href{https://github.com/ZijiaLewisLu/CVPR2025-DeCafNet}{https://github.com/ZijiaLewisLu/CVPR2025-DeCafNet}.
\end{abstract}    

\begin{figure}
    \centering
    % \vspace{-1.5em}
    \includegraphics[width=\linewidth]{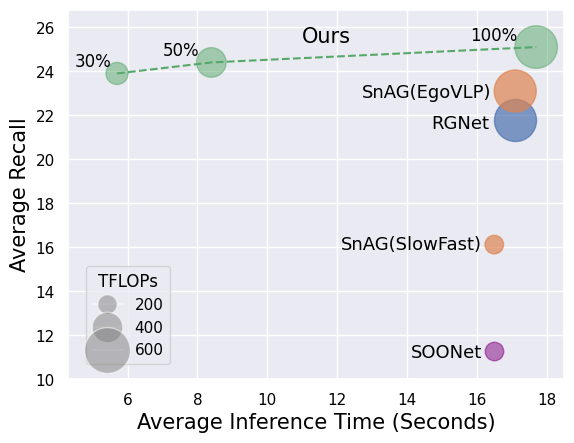}    
    \vspace{-7mm}
    \caption{Model inference time and grounding performance on Ego4d-NLQ \cite{ego4dnlq-hou2023groundnlq} dataset using one A100 gpu. The circle sizes indicate the TFLOPs for methods. Numbers beside the green circles indicate the amount of salient clips processed by expert encoder.}
    \label{fig:intro}
    \vspace{-1.5em}
\end{figure}

\vspace{-4mm}
\section{Introduction}
\label{sec:intro}

% Long Video Temporal Grounding (LVTG) is the task of identifying specific moments or events within long videos (spanning 20 to 120 minutes \cite{}) based on a user-provided text query. Given an input video and text query, the LVTG method aims to pinpoint the exact time segment in the video that matches the text query, allowing efficient retrieval of relevant content from lengthy video sources. 
Long Video Temporal Grounding (LVTG) is the task of identifying specific moments or events within long videos (spanning from several minutes to a few hours \cite{hannanrgnet}) based on a user-provided text query. LVTG allows effective retrieval of relevant content from lengthy videos with a range of applications, such as video summarization \cite{gong2014diverse, potapov2014category, li2023progressive}, content recommendation \cite{khadka2023content,ni2023content}, and surveillance \cite{yu2016long, khor2017lost}, where rapid detection of pertinent segments is critical. %in a time-critical manner.

% State-of-the-art (SOTA) LVTG methods expand on the plethora of works of temporal video grounding in short videos. 

% \textcolor{blue}{state-of-the-art LVTG methods build over techniques developed for short-video temporal video grounding [cite SnAG and RGNet], following a common paradigm -- first dividing long video into clips of fixed duration, processing them through an \textit{expert} encoder, a large, pre-trained model that has been trained on extensive multi-domain video data—to extract video features. Next, they use a localization model with temporal and cross-modal reasoning for the LVTG task. While this paradigm works well for short video, it suffers from scaling challenges on long videos. This is due to the need to process every clip through the LVTG method’s expert encoder imposing a significant computational burden.}

State-of-the-art (SOTA) LVTG methods \cite{hannanrgnet,hou2022cone,pan2023scanning,zhang2020span} build on techniques originally developed for temporal video grounding in short videos, following a common two-step paradigm. First, they divide a long video into fixed-duration clips, processing each clip through an \textit{expert} encoder--a large, pre-trained model trained on extensive, multi-domain video data--to extract video clip features. Second, a grounding model employs temporal and cross-modal reasoning to perform the grounding task. Although effective for short videos, such an approach struggles to scale with longer video lengths due to the high computational cost of processing each clip through the expert encoder. As video duration increases, the number of clips increases, leading to a significant surge in computational demands.

% We  from speculative decoding \cite{leviathan2023fast} in large language models (LLMs) \cite{devlin2019bert, brown2020language} to address the scaling challenge in LVTG. In speculative decoding, not every token requires the same computational effort for accurate generation; similarly, 

Long temporal sequences, whether visual~\cite{cheng2022tallformer}, textual~\cite{chen2023accelerating}, or multimodal~\cite{song2022multimodal}, are computationally heterogeneous, meaning that not all locations in the sequence are equally complex or contribute equally to the final prediction. Consequently, we find that, for a long video, the temporal moment associated with the input query mostly constitutes only a small portion of the whole video. Many clips in a long video are not relevant to the query in LVTG. Therefore, we can process them in a significantly more resource-efficient manner compared to the expert encoder, reducing overall computation required for LVTG and improving scalability while maintaining grounding performance.

Leveraging the mentioned observation, we introduce \textbf{\method} to overcome the limitations of existing methods. \method employs a \textit{delegate-and-conquer} strategy by delegating a significant portion of the computation to an efficient \textit{sidekick} encoder to conquer the computational bottleneck.
The sidekick encoder serves two key functions. First, it computes features for each video clip in a resource-efficient manner to reduce the overall computational cost. Second, it generates a saliency map over video clips by comparing the extracted features with the text query. This allows \method to identify the top-c\% of query-relevant clips that require full-scale processing by the expert encoder. 
% This targeted approach ensures that only the most salient clips undergo intensive computation to achieve high performance. 
This delegate-and-conquer approach ensures that only the most salient clips undergo intensive computation to achieve both high performance and efficiency. %optimizing both efficiency and accuracy.

% To ensure that sidekick encoder determines salient clips effectively, we design two loss functions: Saliency and Distillation losses. 
% These losses jointly optimize the features of the encoder, making it adept at identifying salient regions for full-scale processing.
% These losses jointly optimize the quality of the sidekick's features with minimal computation, making it adept at identifying regions that warrant full-scale processing.

With dense video features extracted by the sidekick encoder for every clip and the top-c\% salient features from the expert encoder, directly using a standard LVTG grounding module like \cite{mu2024snag, hannanrgnet} gives suboptimal performance due to the diverse temporal resolutions of these feature sets. To optimize performance, we introduce \grounding, a novel grounding module designed specifically for our delegate-and-conquer encoder structure. 
% \grounding first aggregates the sidekick and expert video features along a unified temporal scale in a text-query aware manner and then incorporates a multi-scale refinement block to refine the features before performing temporal grounding.
\grounding unifies the features of sidekick and expert encoders via query-aware temporal aggregation and refines them over multiple temporal scales using multi-scale temporal refinement.

We evaluate \method on two LVTG benchmark datasets and successfully achieve stronger grounding performance compared to existing works with $47\%$ less computation on average. 
% Figure~\ref{fig:intro} visualizes \method's computational efficiency with respect to accuracies in Ego4D-NLQ \cite{ego4dnlq-hou2023groundnlq} dataset. 
Figure~\ref{fig:intro} visualizes \method's computational efficiency and accuracy in Ego4D-NLQ \cite{ego4dnlq-hou2023groundnlq} dataset. 
We outperform all existing methods when only processing top-50\% of salient clips using the expert encoder, with negligible additional cost from the sidekick encoder. Overall, \method has 47\% less TFLOPs and 51\% less inference time than the existing works. Our contributions are,
\begin{itemize}

    \item We introduce \method, a novel approach that enhances computational efficiency while improving grounding performance on LVTG.
    \item \method introduces a novel \textit{delegate-and-conquer} approach, employing a combination of a \textit{sidekick} encoder and an \textit{expert} encoder to compute a set of dense and top-c\% salient features. Our \grounding then aggregates and refines these features across multiple temporal scales to perform LTVG optimally.
    % These features are fed to a novel grounding module \grounding that effectively aggregates features at diverse temporal resolution for the grounding task.
    % \item \method combines an efficient sidekick encoder with a query-aware localization model. The sidekick encoder enables low-cost feature extraction, while the localization model ensures flexible, long-range temporal modeling.
 \item \method significantly outperforms existing methods in terms of computational efficiency on two benchmark datasets. Even at much-reduced computation, \method achieves SOTA performance on temporal grounding in long videos, validating both efficiency and effectiveness.  
\end{itemize}

%Ego4D-NLQ and Ego4d-GoalStep

\begin{figure*}[!t]
    \centering
    \includegraphics[width=1.0\linewidth]{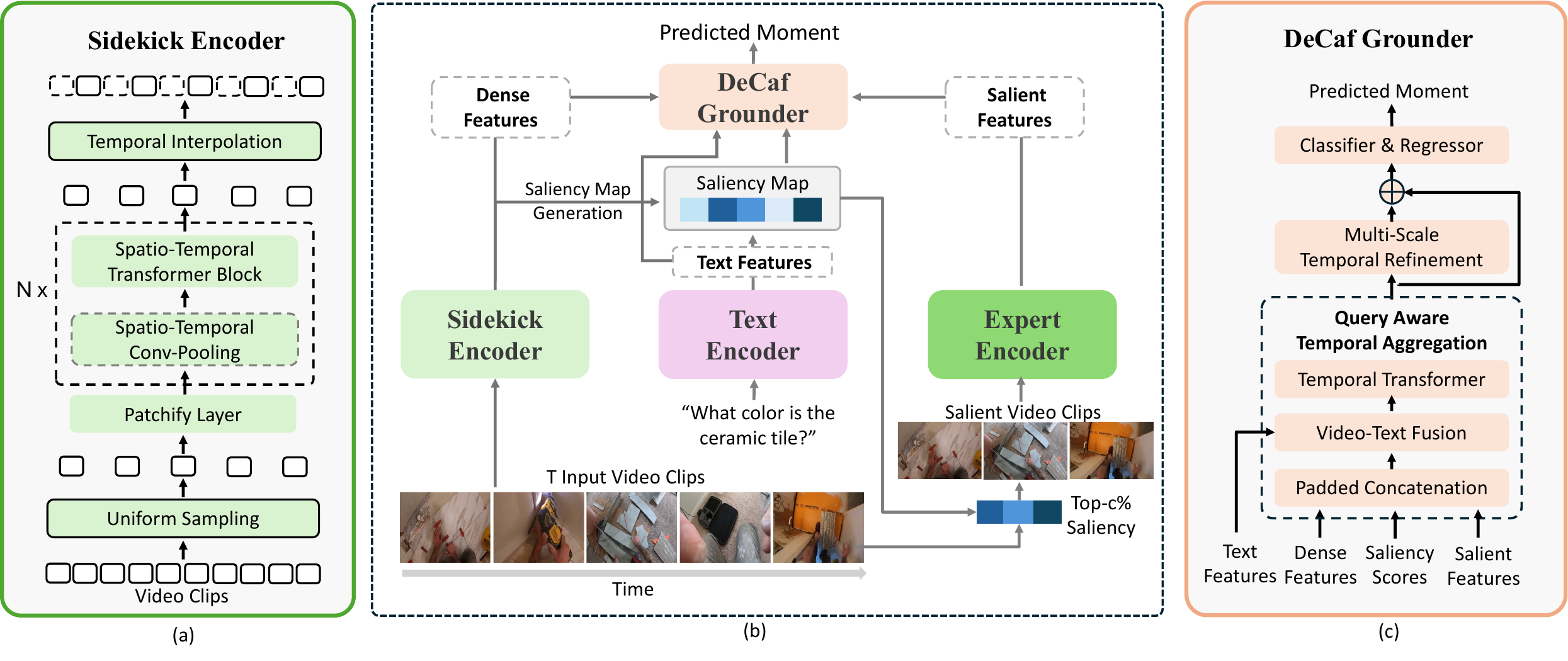}
    \vspace{-2em}
    \caption{\method Overview. The sidekick encoder efficiently extracts features from input video clips, which, combined with text features, generate a saliency map to select the most salient clips. The expert encoder then processes these salient clips. \grounding uses the extracted features from both the sidekick and the expert encoders to predict the moment associated with the input text query.}
    \label{fig:enter-label}
    \vspace{-1em}
\end{figure*} 
\section{Related Works}
\headline{Short Video Temporal Grounding (SVTG).} SVTG methods aim to locate specific events within short videos, typically lasting from a few seconds to a few minutes \cite{regneri-etal-2013-grounding,Sigurdsson2016HollywoodIH,hendricks2017localizingmomentsvideonatural,lei2021detecting}. There is extensive research in this area, which generally falls into proposal-based and proposal-free methods. Earlier proposal-based approaches have used techniques ranging from sliding windows \cite{gao2017talltemporalactivitylocalization,ijcai2018p143,ge2018macminingactivityconcepts,hahn2020trippingtimeefficientlocalization} to ranking mechanisms \cite{wang2019temporallygroundinglanguagequeries,8954090,zeng2020denseregressionnetworkvideo}, to identify candidate segments. Proposal-free methods \cite{Li_2023,woo2022exploreandmatchbridgingproposalbasedproposalfree,pan2023scanning}, on the other hand, leverage Transformer-based algorithms to directly predict start and end points of events. Initial efforts in SVTG detection focused on localizing predefined action categories \cite{wang2021proposalrelationnetworktemporal,zhu2021enrichinglocalglobalcontexts}, meanwhile recent approaches \cite{gao2017talltemporalactivitylocalization,lei2021detecting,pan2023scanning} have broadened their scope by using free-form text queries, such as captions, to locate specific moments. These newer methods explore diverse cross-modal fusion strategies to better align video and textual information for improved grounding performance. However, all SVTG methods face scalability challenges when applied to long video temporal grounding.

\headline{Long Video Temporal Grounding (LVTG)}. 
Long video understanding has been studied for temporal grounding and many other tasks \cite{Lu:CVPR22,Shen:CVPR24,Lee:CVPR24,Donahue:CVPR24,Su:ECCV24,Lu:ICCV21,farha2019ms,zhong2024onlinetas,ding2023every,Ding:CVPR24,Xu_2024_CVPR,reza2022history,Liu_2024_CVPR,Lu_2024_CVPR_MOT}.
% Early LVTG methods expanded SVTG methods with techniques like memory bank, temporal down-sampling and object tracking. 
Early LVTG methods \cite{zhang2020learning,lin2023univtg,zhang2020span,lei2021detecting} expanded SVTG methods with techniques like memory bank \cite{wu2019longtermfeaturebanksdetailed}, sliding window \cite{zhang2020span} and object tracking \cite{wasim2024videogroundingdinoopenvocabularyspatiotemporalvideo}. 
They cannot capture long temporal information, and often struggle to achieve both high efficiency and accuracy. 
More recent methods, such as CONE \cite{hou2022cone}, introduced a coarse-to-fine alignment approach, combining sliding windows, proposal generation, and ranking steps to improve performance. SOONet \cite{pan2023scanning} further refines \cite{hou2022cone} by incorporating pre-ranking and re-ranking techniques to enhance precision. 
Most recently, RGNet \cite{hannanrgnet} approaches LVTG as an integrated retrieval and grounding task, while SnAG \cite{mu2024snag} employs a late fusion strategy to combine textual and visual information in a scalable way. However, most LVTG methods focus primarily on refining the grounding architecture, often overlooking the considerable cost associated with feature extraction for each video clip. These methods typically depend on a pre-trained \textit{expert} encoder for feature extraction. While the cost of such approach is manageable for SVTG, it becomes significantly high for LVTG, where long video lengths amplify the computational burden. Additionally, a substantial portion of the extracted features may not be relevant to the query. Our proposed \method addresses these challenges by delegating a large part of the computation to a more resource-efficient sidekick encoder, reducing unnecessary computational overhead while selecting query-relevant salient clips and processing them with expert encoder, maintaining high grounding performance.
\section{Methodology}

% \subsection{Over}
% \textbf{Overview.}

\subsection{Overview}
Figure~\ref{fig:enter-label} gives an overview of our \method method. Given an input video $V$ and query text $q$, \method aims to localize the temporal moment $(t_s, t_e)$ in the input video that corresponds to the text query. Here, $t_{s}$, $t_{e}$ refer to the start and end timestamp of the moment in the video. 

\method divides the input video $V$ into $T$ fixed duration short clips such that, $V=[v_1, v_2, \ldots, v_T]$. Existing methods \cite{hannanrgnet,mu2024snag, pan2023scanning, zhang2020span} send all the $T$ clips to a pretrained expert encoder, $\Psi_E$, leading to a prohibitive computation requirement particularly for long videos. In contrast, \method adopts a \textit{delegate-and-conquer} strategy to reduce the computational cost. Specifically, we introduce a \textit{sidekick} encoder, $\Psi_D$, that extracts dense clip features, $\F_{D}=[\f'_1, \f'_{2}, \ldots,\f'_{T}]$ at a substantially reduced computational cost. Simultaneously, a text encoder, $\Psi_T$, obtains features $\mathbf{Q}=[\q_\text{cls}, \q_1, \q_2, .., \q_N]$ for the input text query with $N$ number of word tokens and $\q_\text{cls}$ as the CLS token. 

Next, we use $\F_{D}$ and $\mathbf{Q}$ to create a saliency map $\bS$ over the video clips and identify the top-c\% salient clips, corresponding to $M$ ($M$$<$$T$) clips, for the input query. The expert encoder $\Psi_E$ only processes the $M$ salient clips to extract salient features $\F_{S}=[\f_1, \f_{2}, \ldots,\f_{M}]$.

The dense features $\F_{D}$ and the salient features $\F_{S}$ exist at different temporal resolutions. 
% To ensure effective grounding, we introduce a novel grounding module: \grounding to first aggregate them at a unified temporal scale and then refine them over different temporal resolutions using multi-scale refinement. 
To ensure effective grounding, we introduce \grounding that unifies the two features along with the input query features via query-aware temporal aggregation and refines them over varied temporal scales using multi-scale refinement.
Finally, following existing works \cite{hannanrgnet,mu2024snag, pan2023scanning, zhang2020span}, we use regression and classification heads over the refined features to predict the temporal moment ($t_{s}$, $t_{e}$). 
In the following sub-sections, we describe the design of our sidekick encoder, its training, saliency map computation, and \grounding.

\subsection{Sidekick Encoder}

To improve computational efficiency and reduce the number of clips that undergo full-scale processing by the expert encoder $\Psi_E$, we design an efficient sidekick encoder $\Psi_D$ (shown in Figure~\ref{fig:enter-label} (a)) with the following components. 

% \headline{Conv-pooling Operation} 
\headline{Convolution Pooling.} 
$\Psi_D$ follows the architecture of \cite{timesformer}, which is commonly adopted by most modern video encoders.
% \st{Additionally, we add temporal adapter at each layer} \cite{}.}  
It contains a patchify layer with multiple spatio-temporal transformer blocks.
For an input clip $v$, the input to the $i$-th transformer block of $\Psi_D$ is $G_i \in \mathbb{R}^{L\times H\times W\times C}$, where $L, H, W, C$ are the number of frames in $v$, height and width of the feature map, and number of feature channels, respectively. 
To reduce the feature dimension, we insert temporal and spatial pooling layers before the transformer block. We implement this pooling operation through convolutions, with stride size controlling the pooling ratio. By decreasing the temporal ($L$) and spatial ($H, W$) dimensions of the features, we reduce the computational load for the current and subsequent blocks, enabling efficient processing across $\Psi_D$. We determine the value of $i$ empirically. 
% \todo{\st{This pooling strategy is repeated across different layers to maintain lower computational demands as the $\Psi_D$ deepens.} add experiments to it.}

\headline{Temporal Interpolation.} As adjacent clips in a video often contain similar contents, it is possible to infer the features of a clip from nearby clips without having to compute them from scratch. Thus, we further reduce computation for $\Psi_D$ via temporal interpolation. 
Specifically, we first uniformly sample a subset of clips as the input of sidekick encoder and extract a set of features as, $[\f'_1, \f'_{1+\tau}, \f'_{1+2\tau}, ...] = \Psi_D(v_1, v_{1+\tau}, v_{1+2\tau}, ...)$, where $\tau$ is the temporal sampling stride. Then we interpolate the features of the clips that did not get selected during sampling through the extracted features, e.g., 
\begin{align}
    \f'_2, \ldots, \f'_{1+\tau-1} = \operatorname{FFN}([\f'_1, \f'_{1+\tau}]),
\end{align}
where we utilize $(\f'_1, \f'_{1+\tau})$ to interpolate the clips between them, and similarly compute features for other un-sampled clips. Here FFN refers to the feed-forward network. 
% In experiments, we show the effect of different $(\tau, i)$ and their optimal values.

% where we utilize de-convolution for temporal interpolation.
% \todo{
% \begin{align}
%     f'_2, \ldots, f'_{1+s-1} \ldots = \operatorname{FFN}([f'_1, f'_{1+s}]),
% \end{align}
% }

% \zijia
% Given the modification, we are able to fit a whole long video into one A100 GPU and finetune $\Psi_V'$ on LVTG dataset. More importantly, this enables us to learn long temporal information in encoder, which is impossible for full-size encoder as it has to distribute clips of the same video to different GPUs.
% % 
% We add a residual temporal adapter $\Theta$ after $j$-th layer in $\Psi_V'$ to enhance its temporal semantics.

% We finetune $\Psi_V'$ by a saliency loss and distillation loss. Saliency loss improve video-text matching via contrastive learning,
% \begin{align}
%     \mathcal{L}_\text{saliency-text} &= \sum_{e} \frac{\exp(f'_+ \cdot e)}{\exp(f'_+ \cdot e) + \sum_n \exp(f'_- \cdot e)}, \\
%     \mathcal{L}_\text{saliency-video} &= \sum_{f'} \frac{\exp(f' \cdot e_+)}{\exp(f' \cdot e_+) + \sum_n \exp(f' \cdot e_-)},
% \end{align}
% It contains symmetric loss terms for video features and text features. $(f'_+, f'_-)$ are positive and negative clips for a text query obtained from groundtruth LVTG labels, similarly for $(e_+, e_-)$. Distillation loss guide efficient encoder with full-size encoder,
% \begin{align}
%     \mathcal{L}_\text{distill} = || f_t - f_t' ||^2. 
% \end{align}
% \zijiaend

% \ift
\subsection{Sidekick Encoder Training}
To train $\Psi_D$, we introduce two key losses: saliency loss and distillation loss. The saliency loss enhances video-text matching by using contrastive learning~\cite{chen2020simple-contrastive} to align relevant video and text features. This loss includes two symmetrical components for video and text features,
{\small
\begin{align}
    \mathcal{L}_\text{saliency-text} &= \sum_{\cls} \frac{\exp(\f'_+ \cdot \cls)}{\exp(\f'_+ \cdot \cls) + \sum_n \exp(\f'_- \cdot \cls)} \\
    \mathcal{L}_\text{saliency-video} &= \sum_{\f'} \frac{\exp(\f' \cdot \mathbf{q}_{\text{cls}^+})}{\exp(\f' \cdot \mathbf{q}_{\text{cls}^+}) + \sum_n \exp(\f' \cdot \mathbf{q}_{\text{cls}^-})}
\end{align}
}
% where $(\f'_+, \f'_-)$ denote positive and negative video clip features for the given text query, $\cls$ , and $(\mathbf{q}_{\text{cls}^+}, \mathbf{q}_{\text{cls}^-})$ represent positive and negative text features for the given clip feature, $\f'$. Positive and negative clip/text features are obtained from ground-truth LVTG labels. This loss encourages \method to increase the similarity between paired (positive) clips and queries while reducing it for unpaired (negative) instances, enhancing the focus on query-salient video segments.

where $(\f'_+, \f'_-)$ denote positive and negative video clip features that contain and do not contain the queried moment, respectively. $(\mathbf{q}_{\text{cls}^+}, \mathbf{q}_{\text{cls}^-})$ represent text queries that correspond and do not correspond to the given clip feature, $\f'$ respectively. All positive and negative pairs are determined using ground-truth LVTG labels. Saliency loss encourages \method to increase the similarity between paired (positive) clips and queries while reducing it for unpaired (negative) instances, thereby enhancing the focus on query-salient video clips.
%and prevents overfitting to the small-scale training data
The distillation loss further guides $\Psi_D$ by aligning its features with that of the expert encoder, $\Psi_E$. This helps $\Psi_D$ to retain high-quality feature representations. We define the loss as,
\begin{align}
    \mathcal{L}_\text{distill} = || \f_t - \f'_t ||^2. 
\end{align}
Here, $\f_t$ and $\f'_t$ refer to the $t$-th clip features from $\Psi_E$ and $\Psi_D$ respectively. %For each video, since we only pass $M$ clips to $\Psi_E$, the distillation loss is calculated exclusively for these chosen clips.    

\subsection{Saliency Selection} 

% \zijia
% Our efficient encoder processes long videos with much lower computation costs. However, its features are not complete replacement to features from full-size encoder. On the one hand, the pooling layers reduce feature resolutions, thus inevitably lead to information loss and coarse features. On the other hand, as LVTG dataset is smaller than the pretraining datasets, finetuning adapts $\Psi_V'$ to LVTG task but may also slightly reduces its generalization on test data. 
% To harvest the benefit of the both encoder, we proposes a saliency selection methods. Specifically, given a test video, $\Psi_V'$ first process all clips to efficiently obtain coarse clip features $F'$. By comparing $F'$ with the query text feature $e$, a saliency score can be obtained $S\in \mathbb{R}^T=F' \cdot e$, which indicates if a clip is relevant to query text or not. Based on $S$, a subset of the top-$\gamma$\% clips is selected as candidate regions, and passed into full-size encoder to obtain fine-grained features for them. This achieves a balancing between efficiency and accuracy.
% In the end, the final videos features $F^* = \operatorname{concat}[F', \hat{F}, S]$. $\hat{F}$ are features from full-size encoder. $\hat{F}_t = \Psi_V(v_t)$ if t is selected clips and $\hat{F}_t = 0$ otherwise. $S$ is also included to serve as initial localization results to facilitate the localization model.
% \zijiaend

% \ift 
The sidekick encoder, $\Psi_D$ , while efficient, relies on pooling layers that reduce feature resolution, leading to inevitable information loss. 
% While fine-tuning $\Psi_D$ on the smaller LVTG dataset improves alignment with the task, it may reduce generalization.
Therefore, while removing the expert encoder $\Psi_E$ during inference would maximize cost reduction, $\Psi_E$ remains essential for capturing high-quality, detailed features required for LVTG.

% Therefore, we retain $\Psi_E$ but ensure it only processes $M$ numbers of clips among all the $T$ clips for a video. 
We retain $\Psi_E$ but only apply it on the $M$ most salient clips to each query.
We achieve this by creating a saliency map over video clips. Specifically, given an input video, $\Psi_D$ first processes all $T$ clips to obtain dense clip features $\F_D$. By comparing $\F_D$ with the class token of the query text feature $\cls$, we obtain a saliency score through inner product: $\bS=\F_D \cdot \cls \in \mathbb{R}^T$. 
% The saliency score quantifies the semantic saliency between the clip and the text query. 
The saliency score quantifies the semantic relevance between the clips and the text query based on their feature similarity. 
Based on $\bS$, we select top-c\% salient clips (corresponding to $M$ number of clips) and pass them to $\Psi_E$ for extracting salient features $\F_{S}=[\f_1, \f_{2}, \ldots,\f_{M}]$. This delegate-and-conquer dual-encoder design achieves an optimal balance between computational efficiency and feature quality.

\subsection{Grounding Module: \grounding} \label{sec:grounding}
To optimally leverage the features from our delegate-and-conquer dual encoder design, we introduce \grounding to unify and refine the complementary features from the two encoders and locate the temporal moment for text query $\mathbf{Q}$, as shown in Figure \ref{fig:enter-label} (c). 
% 
% The input of \grounding takes two types of features with complias input: 
Dense features $\F_D$ from the sidekick encoder cover every clip in the video while salient features $\F_S$ from the expert encoder have fine-grained semantic information covering the most salient $M$ clips. 
We introduce \textit{Query-aware Temporal Aggregation} that combines $\F_D$ and $\F_S$ to enhance query-relevant information and suppress irrelevant pieces, and \textit{Multi-Scale Temporal Refinement} that efficiently synchronizes information across temporal scales for feature refinement. The features enable \grounding to predict the correct temporal moment $(t_s, t_e)$ for the text query.

% \grounding unifies and refines the dense and salient features and locate the temporal moment for text query $\mathbf{Q}$.

\headline{Query-Aware Temporal Aggregation.} 
% \headline{Feature Alignment and Concatenation} 
% We first perform a concatenation operation between $\F_D$ and $\F_S$. Since $\F_S$ only contains features of salient clips, 
To aggregate $\F_D$ and $\F_S$, comprising features over different sets of clips, we first align their temporal dimensions. 
Specifically, if a clip is missing in $\F_S$, \ie, a non-salient clip, we add zero-padding to its location in $\F_S$ to obtain $\hat{\F}_S$.
With the padding, we ensure $\F_D$ and $\hat{\F}_S$ have the same temporal length, thus allowing a unified feature sequence across all temporal positions.
% \TODO{we first align its temporal dimension with $\F_D$ by padding zeros in the positions where $\F_S$ features are absent.} 
% This zero-padding ensures that both $\F_D$ and $\F_S$ have the same temporal length, allowing for seamless concatenation. The result is a unified feature sequence across all temporal positions. 
We next enhance the text query-specific information by performing video-text fusion to align clip features with text query.
For this, we concatenate $\F_D$ with $\hat{\F}_S$
% , which is the padded salient features that has the same dimensions as $\F_D$.
and further concatenate them with saliency score $\bS$ to provide the explicit context of clip relevance to the text query. This gives us the unified query-aware features $\F_C = \operatorname{concatenate}(\F_D, \hat{\F}_S, \bS) \in \mathbb{R}^{T\times(2C+1)}$. 
Next, we leverage video-text cross-attention that updates $\F_C$ with text query $\mathbf{Q}$ to highlight query-relevant information while suppressing the irrelevant pieces.

% Here $\hat{\F}_S$ represents the padded $\F_S$ that has the same dimensions as $\F_D$.

% To efficiently grounding text query, next we perform video-text fusion to enhance query-related information and suppress irrelevant pieces. Thus we introduce a cross-modal attention block that updates the concatenated features $\F_C$ with query feature $\mathbf{Q}$.
% to suppress query-irrelevant information while maintaining the relevant piece. 

% \headline{Temporal Transformation} 
With the unified query-aware features $\F_C$, we perform temporal aggregation via a temporal transformer~\cite{zhang2022actionformer}. It fuses the information of dense and salient features for each clip while also accounting for the temporal context of neighboring clips.
% It employs local temporal attentions for temporal modeling and 
To capture information at different temporal scales,
it transforms $\F_C$ into multi-scale feature pyramid $\{\Z^l\}_{l=0}^{L}$, where $L$ is the number of scales. 
Each scale reduces the temporal length by half, \eg, $\Z^0 \in \mathbb{R}^{T\times C}$, $\Z^1 \in \mathbb{R}^{T/2\times C}$, and $\Z^l \in \mathbb{R}^{(T/2^{l})\times C}$.
% Similar to prior grounding methods \cite{mu2024snag,pan2023scanning}, each feature in $\{\Z^l\}$ will serve as anchors to predict proposals of temporal moment.

\headline{Multi-Scale Temporal Refinement.} 
While the temporal transformer helps aggregate temporal information, it is confined to only local attention windows. So to learn temporal correlations over longer temporal horizons, we propose multi-scale temporal refinement that efficiently synchronizes grounding-specific information in $\{\Z^l\}$ across temporal scales. 
This is necessary to maintain optimal grounding performance since the features from the two encoders exist at different temporal resolutions.
% In temporal transformer, temporal information is aggregated in local attention windows. 
% 
% To learn temporal co-relations on greater temporal horizons, we propose multi-scale temporal refinement that efficiently synchronizes grounding-specific information in $\{\Z^l\}$ across temporal scales. 
% This is necessary to maintain optimal grounding performance since the features from the two encoders exist at different temporal resolutions.

% We aims to ensure consistent temporal information in $\{\Z^l\}$ across scales. To explicitly focus grounding-specific information and  we synchronize the moment location across scales. 
% Specifically, we aim
%temporal moment happens in the 
Specifically, to explicitly capture grounding information in $\Zl$ (\ie, the probable temporal location of the input text) and reduce feature dimensions, we transform $\{\Z^l\}$ to $\{\mathbf{p}^l\}$ via a simple FFN classifier. $\mathbf{p}^l\in \mathbb{R}^{(T/2^{l})}$ is a confidence score with the same length as $\Zl$. It denotes the probability that a queried moment happens in the temporal locations represented by features in $\Zl$. 
% locate most relevant temporal moment. 
Next, we leverage dilated temporal convolution~\cite{farha2019ms} to synchronize and find the consensus location of the queried moment across scales. This involves first expanding all $\mathbf{p}^l$ to length $T$ via linear interpolation, processing them through convolution, and applying average pooling on output of convolution to obtain a new set of multi-scale features $\{\mathbf{U}^l\}$, which encode refined grounding information.

Lastly, we combine $\Zl$ and $\mathbf{U}^l$ as $\Z^l_\text{refine} = \operatorname{concatenate}(\Z^l, \mathbf{U}^l)$.
$\{\Zl_\text{refine}\}$ unifies the clip features from the sidekick and expert encoders, the text query features, and has highlighted grounding information. 
% More details about multi-scale temporal refinement can be found in the supplementary material.

\headline{Classifier \& Regressor.}
% \headline{Prediction}
% 
% The refined multi-scale features ${Z_{refine}^l}$ are concatenated with ${Z^l}$. 
% 
We use $\{\Zl_\text{refine}\}$ as input to our classification and regression heads to predict proposals of temporal moments. These heads follow the same design as prior works~\cite{mu2024snag, hannanrgnet}. 
At inference, we apply Soft-NMS to merge overlapping moment proposals. 
We use Focal loss and Distance-IoU loss~\cite{mu2024snag} to train \grounding. %the classification and regression heads. %as well as the MLP classifier in multi-scale temporal refinement.

\section{Experiments}

\subsection{Evaluation Settings}

\headline{Datasets.} 
We evaluate our approach on the standard LTVG benchmarks: Ego4D-NLQ, Ego4D-Goalstep and MAD.

\textit{Ego4D-NLQ} \cite{ego4dnlq-hou2023groundnlq}  requires localizing temporal segments~(moments) in videos that contains answer to a natural language query. 
% For example, given a query ``when does the person open the door?", the model needs to identify the precise time interval when the queried content is visible in the video. 
% It contains around $13$ template questions, amounting to around $14K$ queries. 
It contains around $14K$ natural language queries.
% with $11.3K$ and $3.9K$ queries in training and validation sets respectively.
% The training and validation sets contain $11.3K$ and $3.9K$ queries respectively. 
The video length ranges from $8$ to $20$ minutes and the average duration of the temporal moments is $8.3$ seconds. This means the moments constitute only $1.7\%$ of the input videos on average, highlighting the challenge of localizing brief relevant segments within much longer videos.
 %thus requires stronger temporal modeling than Ego4D-NLQ.
\textit{Ego4d-Goalstep} \cite{goldstep-song2023ego4d}  uses action names as text queries. It contains $31.6$K and $7.6$K queries in training and validation sets, respectively. The video length ranges from $1$ to $294$ minutes, with an average of $25$ minutes. The average moment duration is $33$ seconds, constituting only $2.2\%$ of the video on average.
For both Ego4D datasets, as the labels of test sets are unavailable, we follow \cite{mu2024snag,hannanrgnet} and report the performance on validation set. 
\textit{MAD} \cite{soldan2022mad} contains 1.2K hours of movies with 384K queries transcribed from audio description. The videos are 47 to 202 minutes long.

% Some prior works \cite{hannanrgnet,mu2024snag} report performance on the MAD \cite{soldan2022mad} dataset.
% MAD, however, only provides pre-extracted video features without access to the source videos. 
% Since the MAD dataset only provides pre-extracted video features without access to source videos that are needed to train our sidekick encoder, we were not able to evaluate \method on it. 
 
%Thus, we evaluate on the recent Ego4d-Goalstep dataset as an alternative.   
% it computes features from the source videos. 
%Thus, we replace MAD with the recent Ego4d-Goalstep dataset.   
%It prevents us from learning the sidekick encoder. Thus we replace it with Goalstep dataset.

\headline{Metrics.} We adopt the commonly used evaluation metric Recall$@$Top-K with IoU=$\theta$ (denoted as R$k$@$\theta$) \cite{hannanrgnet,mu2024snag}. This metric represents the percentage of test samples with at least one correct prediction among the top-K predictions. A prediction is considered correct if its temporal overlap with the ground truth moment (measured by Intersection over Union) exceeds $\theta$.

\subsection{Implementation Details} 
Similar to \cite{mu2024snag, hannanrgnet}, we partition videos into clips via a sliding window. %, where window length is 32 frames and window stride is 16 frames. 
The input video resolution is 224 $\times$ 224.
In our sidekick encoder, we include the spatio-temporal conv-pooling layer before the first spatio-temporal transformer block, which reduces both spatial and temporal resolutions by a factor of 4. For temporal interpolation, we set $\tau=2$ to process every other clip. To control the balance between contrastive and distillation losses, we set their weights as 1 and 0.75 respectively. In \grounding, we learn multi-scale representations of 8 scales ($L=8$). We use the same expert encoder as in \cite{mu2024snag,kevin2022egovlp,hannanrgnet} and freeze it during training. Please refer to supplementary for more details. 
%We will release our code.

\subsection{Comparison with State-of-the-art.}
\vspace{-1mm}

\begin{table}[]
\resizebox{\linewidth}{!}{%
\fontsize{10pt}{16pt}\selectfont
\setlength{\tabcolsep}{4pt}
\begin{tabular}{l|cc|cc|c}
\textbf{}      & \textbf{R1@0.3} & \textbf{R1@0.5} & \textbf{R5@0.3} & \textbf{R5@0.5} & \multicolumn{1}{c}{\textbf{AVG}} \\ \toprule
2D-TAN \cite{zhang2020learning}  & 5.04  & 2.02  & 12.89 & 5.88  & 6.46  \\
UniVTG \cite{lin2023univtg} & 11.74 & 3.25  & 7.54  & 7.88  & 7.60  \\
VSLNet \cite{zhang2020span} & 10.26 & 5.81  & 19.01 & 12.67 & 11.93 \\
M-DETR \cite{lei2021detecting} & 8.23  & 5.01  & 23.23 & 13.37 & 12.46 \\
SOONet \cite{pan2023scanning}  & 8.00  & 3.76  & 22.40 & 11.09 & 11.31 \\
H-Hands \cite{zhang2023helping} & 13.20 & 7.90  & 23.30 & 15.60 & 15.00 \\
CONE \cite{hou2022cone}    & 14.15 & 8.18  & 30.33 & 18.02 & 17.67 \\
% EgoVLP  & 15.90 & 9.46  & 26.38 & 17.80 & 17.38 \\
% ReLeR   & 19.31 & 11.59 & 23.62 & 15.75 & 17.57 \\
RGNet \cite{hannanrgnet}          & \textit{18.28}           & 12.04           & 34.02           & 22.89           & 21.81                          \\ 
SnAG \cite{mu2024snag}           & 15.87           & 11.26           & \textit{38.26}           & 27.16           & 23.14                          \\
\midrule
% RGNet $\dagger$         & 20.63           & 12.47           & 41.67           & 25.08           & 24.96                          \\ \midrule
\rowcolor{verylightgray}
\method-30\% & 18.07 & \textit{12.41} & 37.68 & \textit{27.47} & \textit{23.91} \\
\rowcolor{verylightgray}
\method-50\%  & \textbf{18.10} & \textbf{12.55} & \textbf{38.85} & \textbf{28.27} & \textbf{24.44}   \\
% \method-100\%~(Ours)  &19.07 & 12.98 & 41.57 & 30.42 & 26.01 \\ \bottomrule     \bottomrule 
% RGNet\cite{hannanrgnet} $\dagger$ & 20.63 & 12.47 & 41.67&  25.08&  24.96 \\
% \method-30\%~(Ours) & 21.13 & 15.04 & 42.42&  31.22&  27.45 \\
% \method-50\%~(Ours) & 20.81 & 15.04 & 42.40 & 31.68 & 27.48 \\
% \method-100\%~(Ours) & 22.21 & 15.52 & 45.63 & 33.93&  29.32 \\ 
\bottomrule 
\end{tabular}
}
\vspace{-2mm}
\caption{Model performance on Ego4D-NLQ dataset. 30\% and 50\% indicate the saliency selection ratio. \method establishes new SOTA with only 50\% saliency clips.}
\label{tab:nlq}
\vspace{-1em}
\end{table}

\begin{table}[]
\centering
\resizebox{0.9\linewidth}{!}{%
\fontsize{10pt}{13pt}\selectfont
\setlength{\tabcolsep}{4pt}
\begin{tabular}{cc|l|l|l}
\multicolumn{1}{c} {$\Psi_D$} & \multicolumn{1}{c|}{$\Psi_E$} & \multicolumn{1}{c|}{TFLOPS} & \multicolumn{1}{c|}{Mem (G)} & \multicolumn{1}{c}{Time (Sec)}  \\ \toprule
100\%    & 0\%        & 21.6                       & 10.9                        & 0.6                                                                               \\
0\%      & 100\%    & 668.2                      & 224.2                       & 17.1                                                                            \\ \midrule
\rowcolor{verylightgray}
100\%    & 30\%     & 222.1  {\scriptsize {\color{Green}$\downarrow$ 66\%}}                       & 79.9  {\scriptsize {\color{Green}$\downarrow$ 65\%}}                        & 5.7 {\scriptsize {\color{Green}$\downarrow$ 67\%}}                                                                               \\
\rowcolor{verylightgray}
100\%    & 50\%     & 355.7 {\scriptsize {\color{Green}$\downarrow$ 47\%}}                      & 126.2 {\scriptsize {\color{Green}$\downarrow$ 44\%}}                       & 8.4 {\scriptsize {\color{Green}$\downarrow$ 51\%}}                                                                               \\ \bottomrule
\end{tabular}
}
\vspace{-2pt}
\caption{Average encoder computation measured on Ego4D-NLQ. 
Column 1, 2 show the amount of clips processed by each encoder.
With saliency selection (row 3, 4), \method significantly reduces TFLOPs by 47\% and 66\% compared to the feature-extraction cost in prior works--processing all clips with expert encoder (row 2).}
\label{tab:computation}
\vspace{-1.6em}
\end{table}

\headline{Ego4d-NLQ \cite{ego4dnlq-hou2023groundnlq}.} We report the model performance on Ego4D-NLQ dataset in Table~\ref{tab:nlq}. We follow prior methods~\cite{mu2024snag,pan2023scanning,hou2022cone} to train \method with only NLQ training data. 
% While \cite{kevin2022egovlp,ramakrishnan2023naq} report results with additional training data and are not comparable, we compare with RGNet \cite{hannanrgnet} using its NLQ-only version for fair evaluation.
% \cite{kevin2022egovlp,ramakrishnan2023naq} report results with additional training data and are not comparable to us. 
For RGNet~\cite{hannanrgnet}, we compare with its NLQ-only version for consistent comparison, and compare with its large-scale pretrained version in supplementary materials.

% \TODO{explain RGNet has a pretraining version and not included here.}
For \method, we evaluate with two ratios: using $\Psi_E$ for only the top 30\% or top 50\% salient clips, %~(Table~\ref{tab:nlq}), 
while processing all clips with our efficient sidekick encoder to achieve a controllable trade-off between computation and accuracy.

\textbf{\method-30\%} uses $\Psi_E$ to process only the top-30\% most salient clips identified by $\Psi_D$. 
% From Table~\ref{tab:nlq}, we can observe that 
Even with this aggressive saliency selection ratio, \method achieves similar or higher performance than prior best method SnAG~\cite{mu2024snag}, improving R1@0.3 and R1@0.5 by 2.3\% and 1.6\% respectively, while being only slightly lower in R5@0.3. It validates that many clips in long videos are not essential for grounding the text query, allowing delegating their computation to our efficient $\Psi_D$ without sacrificing performance.

Next, we report the results of \textbf{\method-50\%}, where $\Psi_E$ processes top-50\% salient clips. This setting consistently outperforms all prior methods across all metrics, exceeding prior works RGNet \cite{hannanrgnet}, and SnAG \cite{mu2024snag} in average recall (AVG) by 2.6\% and 1.3\%, respectively. These results clearly demonstrate the effectiveness of our overall architecture in achieving superior grounding performance while being resource-efficient in computing video clips' features.

\headline{Computation Efficiency.} Having validated the effectiveness of \method, we analyze their computational efficiency in Table~\ref{tab:computation}. To put things into perspective, we first compare the average computation cost of processing the entire video with $\Psi_D$ or $\Psi_E$ (Row 1 vs. Row 2). 
Row 2 also denotes the computation cost of all previous methods, as they use $\Psi_E$ to process 100\% of video clips. 
With the proposed convolution pooling operation and temporal interpolation, $\Psi_D$ achieves a \textbf{31$\times$} reduction in TFLOPs and \textbf{22$\times$} reduction in GPU memory compared to $\Psi_E$. This shows the significantly more efficient design of our sidekick encoder $\Psi_D$ compared to expert encoder $\Psi_E$ and therefore, also compared to all prior methods. 

Thanks to this, Row 3 shows that if we select only top-30\% clips as salient, \method-30\% substantially reduce TFLOPs, GPU Memory, and inference time by 66\%, 65\% and 67\% respectively compared to Row 2. 
If we select top-50\% clips (Row 4), \method-50\% reduces them by 47\%, 44\% and 51\%, respectively compared to Row 2. 
Meanwhile, the TFLOPs of our \grounding is merely 0.06, negligible compared to that of the encoders. 
\method-50\% establishes superior performance (Table~\ref{tab:nlq}) with the significantly lower computational cost~(Table~\ref{tab:computation} Row 4 vs. Row 2). 
This highlights \method's effectivenes in both grounding performance and computational efficiency.
  
% %is on average $46.5\%$ cheaper than the current approaches.  
%%%%%%%%%% OLD %%%%%%%%%%%

\begin{table}[]
\resizebox{\linewidth}{!}{%
\fontsize{9pt}{13pt}\selectfont
\setlength{\tabcolsep}{4pt}
\begin{tabular}{l|cc|cc|c}
\textbf{}    & \textbf{R1@0.3} & \textbf{R1@0.5} & \textbf{R5@0.3} & \textbf{R5@0.5} & \textbf{AVG} \\ \toprule
VSLNet\cite{zhang2020span}      & 11.70            &   -             &  -              &  -              &   -         \\
SnAG\cite{mu2024snag}         & 18.34           & 15.12           & 45.95           & 38.55           & 29.49        
\\ 
RGNet \cite{hannanrgnet}       & \textit{21.26} & 15.71 & \textit{47.15} & \textit{37.85} & \textit{30.49}             \\ \midrule
\rowcolor{verylightgray}
\method-30\%  & 20.01           & \textit{16.22}           & 44.70           & 37.34           & 29.56             \\
\rowcolor{verylightgray}
\method-50\%  & \textbf{21.29}  & \textbf{17.46}  & \textbf{47.27}  & \textbf{40.40}  & \textbf{31.61}    \\ \bottomrule
% \method-100\%~(Ours) & 23.20 & 19.40 & 51.38 & 44.17 & 34.54      \\ \bottomrule
\end{tabular}
}
\vspace{-2mm}
\caption{Model performance on Ego4d-Goalstep dataset. 30\% and 50\% indicating the saliency selection ratio. \method establishes new SOTA with only 50\% saliency clips.}
\label{tab:goalstep}
% \vspace{-0.18in}
\vspace{-0.14in}
\end{table}

\begin{table}[]
\centering
\resizebox{\linewidth}{!}{%
\fontsize{9pt}{13pt}\selectfont
\setlength{\tabcolsep}{4pt}
\begin{tabular}{l|ccc|ccc|c} \toprule
 & \textbf{R1@0.1} & \textbf{R1@0.3} & \textbf{R1@0.5} & \textbf{R5@0.1} & \textbf{R5@0.3} & \textbf{R5@0.5} & \textbf{AVG}\\ \midrule
M-Guide \cite{Barrios2023LocalizingMI} & 9.30 & 4.65 & 2.16 & 18.96 & 13.06 & 7.40 & 9.26 \\
CONE \cite{hou2022cone} & 8.90 & 6.87 & 4.10 & 20.51 & 16.11 & 9.59 & 11.01 \\
SOONet \cite{pan2023scanning} & 11.26 & 9.00 & 5.32 & 23.21 & 19.64 & 13.14 & 13.59 \\
RGNet \cite{hannanrgnet} & \textit{12.43} & \textit{9.48} & \textit{5.61} & \textit{25.12} & 18.72 & 10.86 & 13.70 \\ 
SnAG \cite{mu2024snag} & 10.28 & 8.46 & 5.55 & 24.42 & \textit{20.60} & \textit{13.75} & \textit{13.84} \\ \midrule
\rowcolor{verylightgray}
\method & \textbf{13.25} & \textbf{10.96} & \textbf{7.06} & \textbf{27.73} & \textbf{23.68} & \textbf{16.13} & \textbf{16.47} \\ \bottomrule
\end{tabular}
}
\vspace{-3mm}
\caption{Model performance on MAD dataset. With the same input features, \method outperforms prior works by large margins.}
\vspace{-5mm}
\label{tab:MAD}
\end{table}

\headline{Ego4d-Goalstep~\cite{goldstep-song2023ego4d}}.~We validate \method on Ego4d-Goalstep in Table~\ref{tab:goalstep}. 
% As it is a newer dataset, we replicate the latest prior works RGNet and SnAG on it.
Following our Ego4d-NLQ experiment, we test both 30\% and 50\%  saliency ratios. \method-30\% matches SnAG~\cite{mu2024snag} in average performance, while improving R1@0.3 and R1@0.5 by 2\% and 1\% respectively. \method-50\% demonstrates much stronger performance, outperforming SnAG significantly across all metrics and achieving a 2\% gain in average (AVG) recall. 
% \TODO{}

% For \textbf{Ours (30\%)}, we achieve a similar results as SnAG \cite{mu2024snag} while have higher R1@0.3 and R1@0.5. With 50\% saliency ratio, \textbf{Ours (50\%)}, we outperform SnAG \cite{mu2024snag} and RGNet \cite{hannanrgnet} on average recall (AVG) by 2.12\% and \TODO{}. 
%We like to mention here our evaluation is limited by the 
% Last but not least, the results of \textbf{Ours}$\dagger$ also achieve the best performance on GoalStep, despite videos of this datasets are 3 times longer than those of NLQ dataset on average, demonstrating the robustness of our temporal modeling method.

\headline{MAD~\cite{soldan2022mad}.} 
We evaluate \method on MAD dataset in Table \ref{tab:MAD}. 
MAD only provides pre-extracted video features without the source videos that are needed to train our sidekick encoder. 
Therefore, we train our \grounding with the same input features as all prior methods (i.e., using only features of the expert encoder).
\method successfully outperforms all prior works, exceeding SnAG by 2.63\% in average (AVG) recall. 
% Importantly, we show that our \grounding also achieves better temporal modeling on \textbf{Short Video Temporal Grounding} datasets in Table \ref{tab:charades}, following the settings in \cite{mu2024snag}. We outperform SnAG by 1.37\% and 0.81\% on Charades-STA \cite{Charades} and TACoS \cite{tacos}, respectively. 
% It also highlights the efficacy of our \grounding in isolation of the dual-encoder features and aligns with its performance in \textbf{Short Video Temporal Grounding} datasets. Specifically, in Table \ref{tab:charades}, we evaluate \grounding on Charades-STA \cite{Charades} and TACoS \cite{tacos} where we outperform SnAG by 1.37\% and 0.81\%, respectively. 
It also highlights the efficacy of our \grounding in isolation of the dual-encoder features. We further validate \grounding's performance in \textbf{Short Video Temporal Grounding} datasets. Specifically, in Table \ref{tab:charades}, we evaluate \grounding on Charades-STA \cite{Charades} and TACoS \cite{tacos} where we outperform SnAG by 1.37\% and 0.81\%, respectively.

\begin{table}[]
\centering
\resizebox{0.9\linewidth}{!}{%
\fontsize{9pt}{13pt}\selectfont
\setlength{\tabcolsep}{4pt}
\begin{tabular}{l|cc|cc|c} \toprule
Charades-STA & R1@0.5 & R1@0.7 & R5@0.5 & R5@0.7 & AVG \\ \midrule
% SSRN \cite{} & 45.10 & 34.33 & 65.26 & 51.85 & \\
SMIN \cite{SMIN} & 64.06 & 40.75 & 89.49 & 68.09 & 65.60 \\
SnAG \cite{mu2024snag} & \textit{64.62} & \textit{46.26} & \textbf{92.55} & \textit{71.94} & \textit{68.84} \\
\rowcolor{verylightgray}
\grounding & \textbf{68.79} & \textbf{47.55} & \textit{91.53} & \textbf{72.96} & \textbf{70.21} \\ \midrule\midrule
TACoS & R1@0.3 & R1@0.5 & R5@0.3 & R5@0.5 & AVG \\ \midrule
MATN \cite{MATN} & 48.79 & 37.57 & 67.63 & 57.91 & 52.98 \\
SnAG \cite{mu2024snag} & \textit{56.44} & \textit{44.85} & \textbf{81.15} & \textit{70.66} & \textit{63.27} \\
\rowcolor{verylightgray}
\grounding & \textbf{57.36} & \textbf{46.79} & \textit{81.05} & \textbf{71.13} & \textbf{64.08} \\ \bottomrule
\end{tabular}
}
\vspace{-1mm}
\caption{Model performance on short video temporal grounding datasets. Our \grounding shows better temporal modeling on short videos as well and surpasses SnAG.}
\vspace{-2mm}
% \vspace{-7mm}
\label{tab:charades}
\end{table}

% \begin{table}[]
% \centering
% \begin{tabular}{l|lll|lll}
% \multicolumn{1}{c}{} & \multicolumn{3}{c}{NLQ} & \multicolumn{3}{c}{GoalStep} \\ \toprule
%  & 0.7    & 0.5    & 0.3   & 0.7      & 0.5     & 0.3     \\ \midrule
% sidekick &  90.1   & 80.5  & 66.1     &   \\ 
% fullsize &  89.8   & 80.0  & 65.8     &         \\ \bottomrule
% \end{tabular}
% \caption{Saliency Selection Accuracy.}
% \end{table}

\subsection{Ablation Study}
This section analyzes the effect of different components of \method through ablation. Unless otherwise specified, we evaluate on Ego4d-NLQ using top-50\% salient clips.
% In this section, we analyze the effect of different model components through ablation studies. We report performance on the Ego4d-NLQ dataset using top-50\% salient clips unless specified otherwise.

\begin{table}[]
\resizebox{\linewidth}{!}{%
\fontsize{9pt}{13pt}\selectfont
\setlength{\tabcolsep}{4pt}
\begin{tabular}{ccc|cc|cc|c}
$\F_D$ & $\F_S$ & $\bS$ & \textbf{R1@0.3} & \textbf{R1@0.5} & \textbf{R5@0.3} & \textbf{R5@0.5} & \textbf{AVG} \\ \toprule
$\checkmark$ & $\times$ & $\times$ & 16.32 & 11.32 & 34.08 & 24.33 & 21.51 \\
$\times$ & $\checkmark$ & $\times$ & 16.27 & 11.21 & 36.68 & 26.13 & 22.57 \\
$\checkmark$& $\checkmark$ & $\times$ & 18.12 & 12.84 & 37.11 & 27.16 & 23.91 \\ \midrule
\rowcolor{verylightgray}
$\checkmark$& $\checkmark$ & $\checkmark$ & \textbf{18.87 }  & \textbf{13.18}  & \textbf{38.25}  & \textbf{27.33}  & \textbf{24.41}      \\ \bottomrule
% snag + our features	 17.01	 11.44	 38.28	 27.42	23.5375
\end{tabular}
}
\caption{Effect of encoder features and saliency map. Each feature provides complimentary information and contributes to overall performance.}
\label{tab:saliency}
\vspace{-1em}
\end{table}

\begin{table}[]
\resizebox{\linewidth}{!}{%
\fontsize{9pt}{13pt}\selectfont
\setlength{\tabcolsep}{4pt}
\begin{tabular}{cc|cc|cc|c}
 \begin{tabular}{@{}c@{}}Selection\\Ratio\end{tabular} & 
 \begin{tabular}{@{}c@{}}Selection\\Method\end{tabular} & 
 \textbf{R1@0.3} & \textbf{R1@0.5} & \textbf{R5@0.3} & \textbf{R5@0.5} & \textbf{AVG} \\ \toprule
 30\% & Random& 15.78 & 11.31 & 34.85 & 25.64 & 21.90 \\
 30\% & Uniform& 16.98 & 12.21 & 35.05 & 26.59 & 22.70 \\
 \rowcolor{verylightgray}
 30\% & Saliency  & \textbf{18.21}  & \textbf{12.86} & \textbf{37.48}  & \textbf{27.19} & \textbf{23.94}  \\ \midrule
 
 50\% & Random & 16.98 & 12.04 & 37.39 & 26.93 & 23.33 \\
 50\% & Uniform & 17.15 & 12.44 & 36.94 & 26.84 & 23.34 \\
 \rowcolor{verylightgray}
 50\% & Saliency & \textbf{18.87}  & \textbf{13.18}  & \textbf{38.25}  & \textbf{27.33}  & \textbf{24.41}  \\ \bottomrule
\end{tabular}
}
\caption{Accuracy of saliency maps. Our saliency selection is effective and outperforms both random and uniform selection.}
\label{tab:saliency-acc}
\vspace{-4mm}
\end{table}

%and zero padding elsewhere
\headline{Effect of Encoder Features and Saliency Map.}
Our dual-encoder design provides three features to \grounding: dense features~($\F_D$) from the sidekick encoder, salient features~($\F_S$) from the expert encoder, and the saliency scores $\bS$. Table \ref{tab:saliency} analyzes the contribution of each feature.
First, when using only features $\F_D$ (Row 1, equivalent to selecting top-0\% clips as salient), it leads to lower performance due to inevitable information loss from pooling operations in the sidekick encoder. This indicates the necessity of keeping the high-quality features from the expert encoder. 
% We first show the results of only passing features from the sidekick encoder $\Psi_D$ to grounding module (row 1), which is also equivalent to select top-0\% clips as salient. However, pooling operation leads to inevitable information loss and reduces feature quality, the performance is low. This demonstrates the requirement of keeping the features from the expensive expert encoder, $\Phi_E$. 
%SnAG \cite{mu2024snag} (please see Table \ref{tab:nlq})
Row 2 shows that using only the features $\F_S$, which contains features of just salient clips, also yields lower performance.
% although comparable to prior best method SnAG (Table~\ref{tab:nlq}). 
This is because some features of ground truth moments may be missed due to saliency selection. Row 3 demonstrates that combining both encoders' features provides a balance between recall and efficiency. 
Additionally, incorporating saliency map $\bS$ helps identify candidate regions, further boosting the performance by 0.5\% (row 4).

\headline{Accuracy of Saliency Maps.} 
Table \ref{tab:saliency-acc} compares the performance of random and uniform clip selection against our saliency-based approach.
% It can be observed, when selecting either top-30\% or 50\% salient clips, random selection harms performance, dropping average recall by 0.82\% and 2\% respectively. As expected, the drop is more substantial when fewer clips are selected as salient. This demonstrates our generated salient maps are useful and contain relevant clips.
At both 30\% and 50\% selection ratios, random and uniform selection degrades performance.
As expected, the performance drop is more substantial with the lower 30\% selection ratio, validating the effectiveness of our saliency maps in identifying relevant clips.
Uniform selection outperforms random selection as it evenly samples clips from all regions of the video and is more likely to cover the ground truth moment.
% reducing average recall by 2\% and 1\% respectively. 

\begin{table}[]
\resizebox{\linewidth}{!}{%
\fontsize{9pt}{13pt}\selectfont
\setlength{\tabcolsep}{4pt}
\begin{tabular}{c|cc|cc|cc}
Row  & 
\begin{tabular}[c]{@{}c@{}}Pooling\\ Location ($i$)\end{tabular} & 
\begin{tabular}[c]{@{}c@{}}Temporal\\ Interpolate ($\tau$)\end{tabular} & $\mathcal{L}_\text{saliency}$ & $\mathcal{L}_\text{distill}$ & Recall & TFLOPs \\ \toprule
\rowcolor{verylightgray}
1& 1 & 2 & $\checkmark$ & $\checkmark$ & 80.5 & 21.6 \\ \midrule
2& 2 & 2 & $\checkmark$ & $\checkmark$ & 81.0 & 38.9 \\
3&3 & 2 & $\checkmark$ & $\checkmark$ & 81.5 & 53.9 \\ \midrule
4&1 & 1 & $\checkmark$ & $\checkmark$ & 82.6 & 42.7 \\
5&1 & 3 & $\checkmark$ & $\checkmark$ & 72.8 & 14.4 \\ \midrule
6&1 & 2 & $\times$ & $\checkmark$ & 51.1 & 21.6 \\
7&1 & 2 & $\checkmark$ & $\times$ & 48.5 & 21.6 \\ \bottomrule
\end{tabular}

}
\vspace{-2mm}
\caption{Ablation study on sidekick encoder. Our default parameters (row 1) strike a balance between accuracy and efficiency. Here, recall refers to the percentage of ground truth clips present within the top 50\% salient clips. }
\vspace{-1.3em}
\label{table:ablation-sidekick}
\end{table}

\begin{table}[]
\resizebox{\linewidth}{!}{%
\fontsize{9pt}{13pt}\selectfont
\setlength{\tabcolsep}{4pt}
\begin{tabular}{cc|cc|cc|c}
QTA & MTR & \textbf{R1@0.3} & \textbf{R1@0.5} & \textbf{R5@0.3} & \textbf{R5@0.5} & \textbf{AVG} \\ \toprule
$\times$ & $\checkmark$ & 16.34 & 11.21 & 36.13 & 25.79 & 22.37 \\ 
$\checkmark$ & $\times$ & 16.41 & 11.58 & 37.62 & 27.13 & 23.18 \\ \midrule
\rowcolor{verylightgray}
$\checkmark$& $\checkmark$ & \textbf{18.87} & \textbf{13.18}  & \textbf{38.25}  & \textbf{27.33}  & \textbf{24.41} 
\\ \bottomrule
% Ours (100\%) & 21.87           & 18.41           & 51.27           & 43.74           & 33.82        \\ \bottomrule
\end{tabular}
}
\vspace{-2mm}
\caption{Ablation for \grounding. QTA and MTR stand for Query-aware Temporal Aggregation and Multi-scale Temporal Refinement. Both contribute significantly to the performance.}
\label{tab:grounding}
\vspace{-3.5mm}
\end{table}

\begin{table}[]
\centering
\resizebox{0.95\linewidth}{!}{%
\fontsize{9pt}{13pt}\selectfont
\setlength{\tabcolsep}{4pt}
\begin{tabular}{c|cc|cc|c}
 & \textbf{R1@0.3} & \textbf{R1@0.5} & \textbf{R5@0.3} & \textbf{R5@0.5} & \textbf{AVG} \\ \toprule
SnAG\cite{mu2024snag}           & 15.87           & 11.26           & 38.26           & 27.16           & 23.14                          \\
% Ours $\dagger$ & 16.41 & 11.58 & 37.62 & 27.13 & 23.18 \\ \midrule
\rowcolor{verylightgray}
\grounding & \textbf{19.07}   & \textbf{13.61}  & \textbf{39.02}  & \textbf{29.22}  & \textbf{25.10} \\ \bottomrule      
% Ours (100\%) & 21.87           & 18.41           & 51.27           & 43.74           & 33.82        \\ \bottomrule
\end{tabular}
}
\vspace{-2mm}
\caption{Grounding performance with same input features as prior art \cite{mu2024snag}. \grounding outperforms on all metrics.}
% This setting is same as using 100\% features from the expert encoder with no features from the sidekick.}
\label{tab:grounding2}
\vspace{-1.2em}
\end{table}

\begin{table}[]
\centering
\resizebox{0.95\linewidth}{!}{%
\fontsize{9pt}{13pt}\selectfont
\setlength{\tabcolsep}{4pt}
\begin{tabular}{c|cc|cc|c}
 & \textbf{R1@0.3} & \textbf{R1@0.5} & \textbf{R5@0.3} & \textbf{R5@0.5} & \textbf{AVG} \\ \toprule
% No-Augment & 18.87 & 13.18  & 38.25  & 27.33  & 24.41 \\ \midrule
Lighting     & 18.27 & 12.64 & 38.14 & 27.44 & 24.12  \\
Blurring & 17.47 & 11.86 & 36.88 & 26.53 & 23.18  \\       
Occlusion & 18.04 & 12.41 & 37.76 & 26.82 & 23.75 \\ \bottomrule
\end{tabular}
}
\vspace{-3mm}
\caption{Ablation on model robustness with data augmentation to lower video lighting, increase blurring and add occlusion.}
\label{tab:robust}
\vspace{-2em}
\end{table}

\begin{figure*}[t]
    \centering
    \includegraphics[width=0.9\linewidth,]{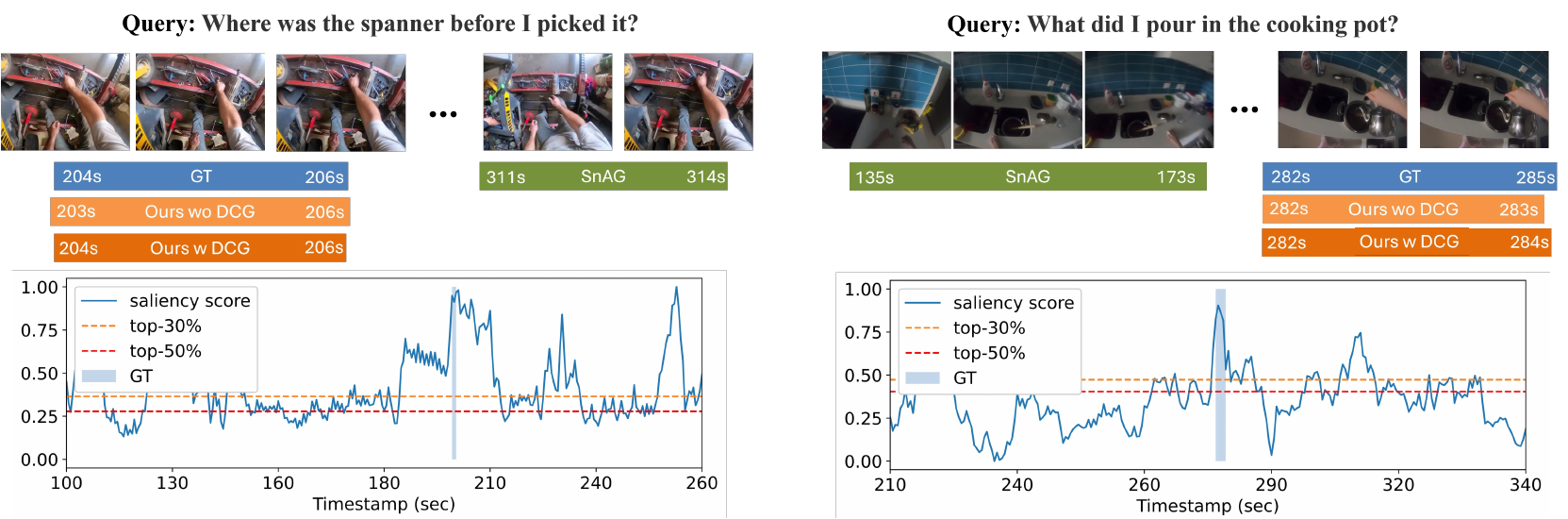}
    \vspace{-3mm}
    \caption{\method's qualitative results, where Ours wo DCG and Ours w DCG indicate predictions without and with \grounding, respectively. The second row displays the generated saliency maps.
    \method yields accurate saliency maps and better grounding results.
    % Our superior performances compared to SnAG \cite{mu2024snag} can be seen in all examples. 
    % \method's saliency maps are accurate, frequently capturing ground truth information even with significantly reduced computation of using only the top 30\% of salient clips. This highlights the effectiveness of our dual encoder design. 
    % \todo{For Ours wo DCG, we use Snag's \cite{mu2024snag} grounding module. Traditional grounding modules, which lack diverse temporal resolution consideration, often misalign the query (as shown in row-2), whereas our \grounding effectively leverages these features for accurate grounding.}  
    }
        \vspace{-5mm}
    \label{fig:qual}
\end{figure*}

% \zijia
% The following are running experiments for encoder ablation study.
% \zijiaend

% Effect 

\headline{Effect of Sidekick Encoder.}
% Table \ref{table:ablation-sidekick} evaluates the efficiency and saliency selection accuracy of our sidekick encoder.
Table \ref{table:ablation-sidekick} shows the effect of different design choices of convolution-pooling and temporal interpolation. 
We use recall to measure the feature quality of sidekick encoder -- the success rate of including ground truth clips when selecting top 50\% salient clips.
% by ranking the relevance of video clips to a query based on saliency score $S$ and compute mAP.
Row 1 shows our default configuration: spatial/temporal pooling before the first transformer block $(i = 1)$ and temporal interpolation ratio $\tau = 2$. This achieves 80\% recall, meaning our efficient sidekick encoder successfully identifies most of ground truth clips for expert processing. 
% Figure \ref{fig:qual} (row 2) also demonstrates this strong performance.

Next, adding pooling in later blocks (row 2-3) slightly improves recall but substantially increases computation costs. Thus we set $i=1$ to maintain efficiency. Similarly, 
varying the temporal interpolation ratio $\tau$ (row 4-5) shows $\tau=2$ strikes a balance between accuracy and efficiency.

\headline{Effect of Encoder Losses.} The bottom section of Table \ref{table:ablation-sidekick} (row 6-7) demonstrates the importance of both saliency loss and distillation loss. Removing $\mathcal{L}_\text{saliency}$ prevents the sidekick encoder from learning accurate video-text similarities, while removing $\mathcal{L}_\text{distill}$ limits the encoder's ability to learn generalizable features from limited training data. Both cases result in significant recall degradation.

\headline{Effect of \grounding.} Table~\ref{tab:grounding} studies the effect of the two key components in \grounding: query-aware temporal aggregation (QTA) and multi-scale temporal refinement (MTR). %by removing either query-aware temporal aggregation (QTA) or multi-scale temporal refinement (MTR). 
\grounding receives features from both encoders: dense features from $\Psi_D$ covering all clips and salient features from $\Psi_E$ covering only salient clips. 
% Effective temporal aggregation of these complementary features is crucial for optimal performance. 
%Effective temporal aggregation of these complementary features is crucial for optimal performance.
% As mentioned, the inputs to the grounding module contain features of both encoders, where features from $\Psi_D$ cover all clips but missing some granular details and features of $\Psi_E$ contain all details yet only covers a subset of salient frames. It is important for grounding module to aggregate their temporal information. %to leverage the benefit of both features. 
Removing QTA~(Row 1) results in a 2.04\% reduction in average recall, highlighting the importance of guiding feature fusion via text query.
% QTA in effective feature aggregation.
Similarly, removing MTR~(Row 2) decreases average recall by 1.23\%, showing the negative impact of limiting temporal modeling to local attention windows. 
Table~\ref{tab:grounding2} evaluates \grounding by using the same input features as SnAG~(equivalent to 100\% feature computation via $\Psi_E$). This setting outperforms SnAG by 1.9\% on average recall, further proving the efficacy of \grounding. 

\headline{Model Robustness.}
% To evaluate our model robustness, we test 
In Table \ref{tab:robust}, we evaluate our model robustness on three challenging scenarios by applying data augmentations on Ego4D-NLQ test videos. These include: (1) reduce lighting by 50\%, (2) apply Gaussian blurring\footnote{we use kernel size=(7, 11), standard deviation = (0.1, 5)}, and (3) mask 10\% of pixels for occlusion. Our average recall drops slightly from 24.41 to 24.12, 23.18, and 23.75 in three settings, yet remains \textit{higher than prior best result with no augmentation} (SnAG - 23.14), validating our robustness.

\vspace{-1mm}
\subsection{Qualitative Results}
\vspace{-2mm}
% Figure \ref{fig:qual} shows qualitative results from our model. In bottom we show saliency maps. Ours (Initial) and Ours (Final) refer to the predictions without and with \grounding respectively. We compare our predictions with SOTA model\cite{mu2024snag}. We can see \method's saliency maps are accurate and cover the ground truths. Even if we use top 30\% salient clips the ground truths lie within them. This shows how effectively our dual-encoder design can reduce computation while maintaining accuracy. In case of Ours (Initial), we utilize Snag's \cite{mu2024snag} grounding module. As input we use the concatenated features as described in \ref{sec:grounding}. It can be seen without \grounding the grounding predictions are not always correct (Figure \ref{fig:qual} (b)). This is expected as traditional grounding models do not consider input at diverse temporal resolutions. However, this failure cases can be fixed with \grounding.  

Figure \ref{fig:qual} presents qualitative results of our model, with saliency maps displayed at the bottom of our predictions. Ours wo DCG and Ours w DCG represent predictions without and with \grounding, respectively. 
In Ours wo DCG, we employ SnAG’s grounding module, using concatenated features as input. The concatenation of features is described in section \ref{sec:grounding}. 
% We compare our predictions with SnAG. 
Notably, \method’s saliency maps are accurate and consistently align with the ground truth. Even when only the top 30\% of salient clips are considered, they still capture the ground truth, demonstrating the effectiveness of our dual-encoder design. 
% As shown in Figure \ref{fig:qual} row 2, 
Moreover, predictions without \grounding are occasionally inaccurate, as existing grounding models do not consider inputs of different temporal resolutions. However, these cases are effectively corrected with \grounding.

% \begin{figure}
%     \centering
%     \includegraphics[width=\linewidth]{Figures/qualitative3/q1.pdf}
%     \caption{Caption}
%     \label{fig:res1}
% \end{figure}

% \begin{figure}
%     \centering
%     \includegraphics[width=\linewidth]{Figures/qualitative3/q2.pdf}
%     \caption{Caption}
%     \label{fig:res2}
% \end{figure}

% \begin{figure}
%     \centering
%     \includegraphics[width=\linewidth]{Figures/qualitative3/q3.pdf}
%     \caption{Caption}
%     \label{fig:res2}
% \end{figure}

% \begin{figure}
%     \centering
%     \includegraphics[width=\linewidth]{Figures/qualitative/2b19ee03-d94f-4c1f-822d-d4c00a2066cc-4.jpg}
%     \caption{Caption}
%     \label{fig:enter-label}
% \end{figure}

% \begin{figure*}[h!]
%     \centering
%     \begin{subfigure}[t]{0.3\linewidth}
%         \centering
%         \includegraphics[width=\linewidth]{Figures/qualitative3/q1.pdf}
%         \caption{Caption for first figure}
%         \label{fig:res1}
%     \end{subfigure}%
%     \hfill
%     \begin{subfigure}[t]{0.3\linewidth}
%         \centering
%         \includegraphics[width=\linewidth]{Figures/qualitative3/q2.pdf}
%         \caption{Caption for second figure}
%         \label{fig:res2}
%     \end{subfigure}%
%     \begin{subfigure}[t]{0.3\linewidth}
%         \centering
%         \includegraphics[width=\linewidth]{Figures/qualitative3/q3.pdf}
%         \caption{Caption for third figure}
%         \label{fig:res3}
%     \end{subfigure}
%     \caption{Overall caption for the combined figure}
%     \label{fig:combined}
% \end{figure*}

\vspace{-3mm}
\section{Conclusion}
\vspace{-2mm}
We present \method, a novel approach that employs a \textit{delegate-and-conquer} strategy to improve computational efficiency without sacrificing performance for LVTG task. 
\method introduces a sidekick encoder that efficiently computes features for all clips, while generating a saliency map to identify the most salient clips for full processing by the expert encoder. 
% This strategy reduces computational load by only focusing intensive computation on the most salient video segments. 
We validate \method on three datasets. It achieves SOTA grounding performance while reducing computation by up to 47\%, making it a promising solution for LVTG tasks.
% in terms of both computational efficiency and grounding accuracy. DelNet reduces computation by up to 47\%, while still achieving state-of-the-art performance in long video temporal grounding tasks. 
% Our results highlight \method can balances efficiency and effectiveness, 

{
    \small
    \bibliographystyle{ieeenat_fullname}
    \bibliography{main}
}

\clearpage
\setcounter{page}{1}
\maketitlesupplementary

% - Explain multi-scale refinement
% - Explain we replicate the results on goalstep
% - run experiment with Uniform sampling

We present \method, an efficient algorithm that uses a \textit{delegate-and-conquer} strategy to achieve accurate and efficient temporal grounding in long videos. In this supplementary material, we provide additional details about our architecture, experimental results, ablation studies, and implementation specifics. 
\section{Additional Architectural Details}
\begin{figure*}
    \centering
    \includegraphics[width=0.8\linewidth]{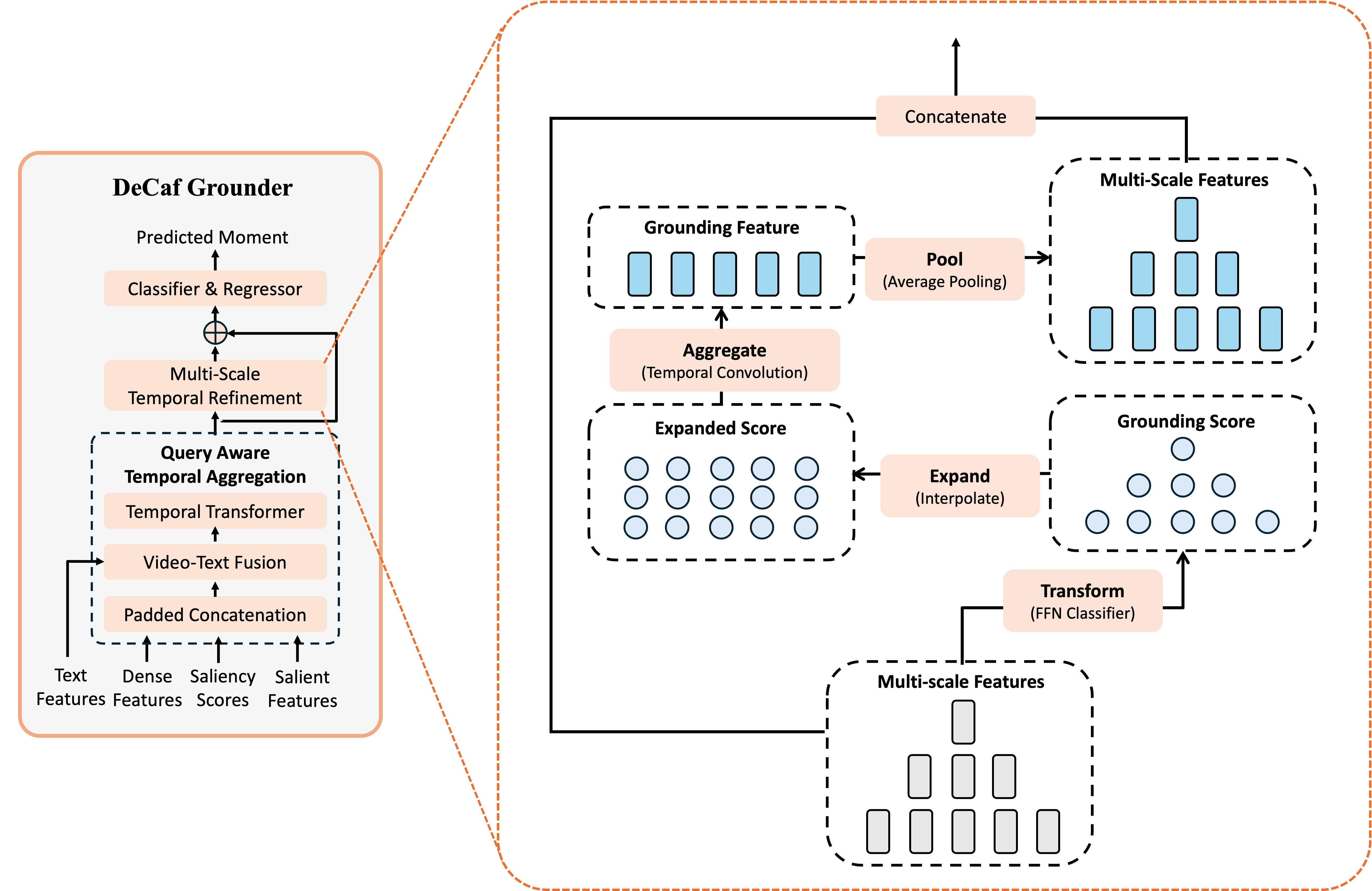}
    \caption{Details of multi-scale temporal refinement. The multi-scale features produced by the temporal transformer are transformed into grounding scores using an FFN classifier. To synchronize grounding information across different scales, we utilize linear interpolation and temporal convolution. Finally, average pooling is applied to effectively combine the synchronized features with the input features. }
        \label{fig:mstr}
    \vspace{-3mm}
\end{figure*}
% In this section we provide detailed explanation for our multi-scale temporal refinement in \grounding.

To enable temporal grounding using features extracted by both the sidekick and expert encoders, we introduce \grounding. \grounding consists of the following key components: query-aware temporal aggregation, multi-scale temporal refinement, and classifier \& regressor. In this section, we provide additional details about the multi-scale temporal refinement component.   

Recall that, \grounding produces multi-scale features via query-aware temporal aggregation, $\{\Zl\}_{l=0}^L$. The features capture temporal information from local to global scales, i.e., $\Z^0$ represents the most local scale, encoding one clip per feature, and $\Z^L$ is the most global scale, encoding $2^L$ clips per feature.
% \TODO{E.g., $\Z_0$ represents the most-local scale, with one feature per clip. $\Z_1$ has }.
%Since dense features $\F_D$, from the sidekick encoder and salient features $\F_S$, from the expert encoder exist at different temporal resolutions

Since the features generated by the sidekick and expert encoders are at different temporal resolutions, this mismatch can result in inconsistencies in $\Zl$ across varying scales. We aggregate information across scales to improve temporal grounding and focus on grounding-relevant information to maximize efficiency.
Overall, multi-scale temporal refinement consists of four steps: \textbf{transform-expand-aggregate-pool}, as shown in Figure \ref{fig:mstr}.
% we explain each step one by one.

\headline{Transform.} To explicitly capture grounding-specific information, we transform $\{\Zl\}$ to $\{\bpl\}$ via a FFN classifier, 
\begin{align}
    \bpl = \operatorname{FFN}(\Zl) \in \mathbb{R}^{T/2^l}.
\end{align}
where $\bpl$ has the same temporal length as $\Zl$. It explicitly denotes if the ground truth moments happen at the temporal position represented by features in $\Zl$. 
This also reduces the feature dimension to 1.
The FFN classifier is trained via Focal Loss as explained in the main paper. 
%, improving efficiency
% \TODO{For example, let $\bpl_i$ be its $i$-th entry. A high $\bpl_i$ means a high likelihood that the moment happens around timestamp $i T/2^l$.}

\headline{Expand.} To combine $\{\bpl\}$, we need to first align their temporal lengths. We apply linear interpolation to expand each $\bpl$ to length $T$,
\begin{align}
    \bhpl = \operatorname{linear-interpolate}(\bpl) \in \mathbb{R}^T.
\end{align}
All $\{\bhpl\}$ have the same temporal length $T$. Thus, we can concatenate them to obtain $\bhP = \operatorname{concat}[\hat{\mathbf{p}}^0, \ldots, \hat{\mathbf{p}}^T] \in \mathbb{R}^{T\times L}$.

\headline{Aggregate.} With $\bhP$, we employ a temporal convolution to synchronize grounding information across scales,
\begin{align}
    \bH = \operatorname{convolution}(\bhP) \in \mathbb{R}^{T \times C},
\end{align}
where $\bH$ is the output of temporal convolution, encoding refined grounding information. $C$ is the size of feature dimension.

\headline{Pool.} To combine $\bH$ with the initial features $\{\Zl\}$, we continue to compute a multi-scale feature pyramid from $\bH$ via simple average pooling, 
\begin{align}
    \bUl = \operatorname{average-pooling}(\bH) \in \mathbb{R}^{T/2^l \times C},
\end{align}
where $\bUl$ is obtained by pooling $\bH$ on temporal dimension by a factor of $2^l$. Finally, we concatenate it with $\Zl$ to obtain $\Zl_\text{refine}$ as explained in the main paper.

\section{Computation Efficiency on Ego4D-Goalstep}
\begin{table}[]
\centering
\resizebox{0.9\linewidth}{!}{%
\fontsize{10pt}{13pt}\selectfont
\setlength{\tabcolsep}{4pt}
\begin{tabular}{cc|l|l|l}
\multicolumn{1}{c} {$\Psi_D$} & \multicolumn{1}{c|}{$\Psi_E$} & \multicolumn{1}{c|}{TFLOPS} & \multicolumn{1}{c|}{Mem (G)} & \multicolumn{1}{c}{Time (Sec)}  \\ \toprule
100\%    & 0\%        & 64.8 & 40.1  & 1.9  \\
0\%      & 100\%    & 2071.8 & 700.4  & 48.0 \\ \midrule
\rowcolor{verylightgray}
100\%    & 30\%     & 686.3  {\scriptsize {\color{Green}$\downarrow$ 67\%}}                       & 250.2  {\scriptsize {\color{Green}$\downarrow$ 64\%}}                        & 15.3 {\scriptsize {\color{Green}$\downarrow$ 68\%}}                                                                               \\
\rowcolor{verylightgray}
100\%    & 50\%     & 1100.7 {\scriptsize {\color{Green}$\downarrow$ 47\%}}                      & 390.3 {\scriptsize {\color{Green}$\downarrow$ 44\%}}                       & 24.3 {\scriptsize {\color{Green}$\downarrow$ 49\%}}                                                                               \\ \bottomrule
\end{tabular}
}
\caption{Average Encoder Computation measured on Ego4D-GoalStep \cite{goldstep-song2023ego4d} dataset. 
Column 1, 2 show the amount of clips processed by each encoder.
With saliency selection (row 3, 4), \method significantly reduces TFLOPs by 47\% and 67\% compared to the feature-extraction cost in prior works that process all clips with expert encoder $\Psi_E$ (row 2).}
\label{tab:computation2}
\end{table}

In Table 2 of the main paper, we have reported computation efficiency on Ego4D-NLQ dataset. In Table \ref{tab:computation2} of this supplementary material, we also show the computation on Ego4D-Goalstep dataset.
Row 2 shows the feature extraction cost of all prior works that process all clips via expert encoder $\Psi_E$. Row 3 and 4 show the computation of our saliency selection method with the sidekick encoder $\Psi_D$. Since the computation cost is linear to the number of video clips, we similarly reduce TFLOPS by 67\% and 47\%, demonstrating our \textit{delegate-and-conquer} approach has significantly lower computation cost than prior methods.

\begin{table}[]
\resizebox{\linewidth}{!}{%
\fontsize{10pt}{16pt}\selectfont
\setlength{\tabcolsep}{4pt}
\begin{tabular}{l|cc|cc|c}
\textbf{}      & \textbf{R1@0.3} & \textbf{R1@0.5} & \textbf{R5@0.3} & \textbf{R5@0.5} & \multicolumn{1}{c}{\textbf{AVG}} \\ \toprule
RGNet \cite{hannanrgnet}          & 18.28           & 12.04           & 34.02           & 22.89           & 21.81                          \\ 
SnAG \cite{mu2024snag}           & 15.87           & 11.26           & 38.26           & 27.16           & 23.14                          \\
\midrule
\rowcolor{verylightgray}
\method-30\% & 18.07 & 12.41 & 37.68 & 27.47 & 23.91 \\
\rowcolor{verylightgray}
\method-50\%  & \textbf{18.10} & \textbf{12.55} & \textbf{38.85} & \textbf{28.27} & \textbf{24.44}   \\
\rowcolor{beaublue}
\method-100\% &\textbf{19.07} & \textbf{12.98} & \textbf{41.57} & \textbf{30.42} & \textbf{26.01} \\ \bottomrule     
\bottomrule 
RGNet\cite{hannanrgnet} $\dagger$ & 20.63 & 12.47 & 41.67&  25.08&  24.96 \\
\rowcolor{verylightgray}
\method-30\% $\dagger$& \textbf{21.13} & 15.04 & \textbf{42.42} &  31.22&  27.45 \\
\rowcolor{verylightgray}
\method-50\% $\dagger$ & 20.81 & \textbf{15.04} & 42.40 & \textbf{31.68} & \textbf{27.48} \\
\rowcolor{beaublue}
\method-100\% $\dagger$ & \textbf{22.21} & \textbf{15.52} & \textbf{45.63} & \textbf{33.93} & \textbf{29.32} \\ \bottomrule 
\end{tabular}
}
\caption{Complete Model Results on Ego4D-NLQ dataset. $\dagger$ denotes the models are pretrained on NaQ dataset \cite{ramakrishnan2023naq}.}
\label{tab:nlq}
\end{table}

\begin{table}[]
\resizebox{\linewidth}{!}{%
\fontsize{9pt}{13pt}\selectfont
\setlength{\tabcolsep}{4pt}
\begin{tabular}{l|cc|cc|c}
\textbf{}    & \textbf{R1@0.3} & \textbf{R1@0.5} & \textbf{R5@0.3} & \textbf{R5@0.5} & \textbf{AVG} \\ \toprule
VSLNet \cite{zhang2020span}      & 11.70            &   -             &  -              &  -              &   -         \\
SnAG \cite{mu2024snag}         & 18.34           & 15.12           & 45.95           & 38.55           & 29.49        
\\ 
RGNet \cite{hannanrgnet}       & 21.26 & 15.71 & 47.15 & 37.85 & 30.49             \\ \midrule
\rowcolor{verylightgray}
\method-30\%  & 20.01           & 16.22           & 44.70           & 37.34           & 29.56             \\
\rowcolor{verylightgray}
\method-50\%  & \textbf{21.29}  & \textbf{17.46}  & \textbf{47.27}  & \textbf{40.40}  & \textbf{31.61}    \\ 
\rowcolor{beaublue}
\method-100\% & \textbf{23.20} & \textbf{19.40} & \textbf{51.38} & \textbf{44.17} & \textbf{34.54}      \\ \bottomrule
\end{tabular}
}
\vspace{-2mm}
\caption{Complete Model Results on Ego4D-Goalstep dataset.}
\label{tab:goalstep}
\vspace{-0.14in}
\end{table}

\section{Additional Experimental Results}
Table \ref{tab:nlq},~\ref{tab:goalstep} show complete model results on Ego4D-NLQ and Ego4D-Goalstep datasets.
Their settings are consistent with those of Table 1, 3 in the main paper.
We include the performance of \method-100\% on both datasets, where we process all clips with both sidekick and expert encoders (rows in blue in Table \ref{tab:nlq} and Table \ref{tab:goalstep}). 
Compared to all prior methods that process all clips with the expert encoder, this model provides more diverse features to grounding models with \textit{only 3\% more TFLOPs} for running the sidekick encoder (row 1 vs row 2 in Table \ref{tab:computation2}).
It can be observed that, \method-100\% greatly boosts the performance. In Table \ref{tab:nlq}, it achieves an average recall of 26.01\% on Ego4D-NLQ, higher than SnAG by 2.87\%. In Table \ref{tab:goalstep}, it achieves an average recall of 34.54\% on Ego4D-Goalstep, higher than SnAG by 4.05\%.
% \vspace{-2mm}

Moreover, we also follow the setting in RGNet to pretrain models on the larger NaQ dataset \cite{ramakrishnan2023naq}, as shown in the second section of Table \ref{tab:nlq}.
First, we highlight that, our \method-50\% without pretraining already achieves close performance to RGNet with pretraining, while using 47\% less computations. After pretraining, \method outperforms RGNet by large margins and improves average recall by 2.49\% to 4.36\%. Pretraining also enhances our accuracy on saliency selection, therefore \method-30\% now has similar performance as \method-50\%.

\section{Implementation Details}
Our sidekick encoder has 12 spatio-temporal blocks and we initialize its weight from \cite{kevin2022egovlp} to speed up training.
For temporal convolution \cite{Lu:CVPR24} in multi-scale temporal refinement, we use 8 layers, where the dilation rate of the $i$-th convolution layer equals to $2^i$.
% \TODO{think of more details to add.}
Since neither SnAG nor RGNet reports performance on the Ego4D-Goalstep dataset, we use their released codes to report performance on this dataset.
We measure all computation cost using one 80GB A100 GPU. When the GPU cannot store all video clips in memory, we split the data into multiple batches and report the overall TFLOPS/Mem/Time summed over all batches.
To evaluate on short temporal video grounding datasets, we use features released by SnAG and use the I3D feature for Charades-STA dataset.

\section{Limitations}

% While our \method approach has established new SOTA for LVTG with greatly reduced computation, the overall grounding performance of all methods so far are relatively low, especially for R1@0.3 and R1@0.5. 

\method has established new SOTA for LVTG with greatly reduced computation. However, the overall recall values are relatively low, especially for R1@0.3 and R1@0.5. We found this is partly caused by ambiguity in text queries in the dataset. For example, for a text query of ``Where was object X before I used it?'', the object was often used for multiple times by the person. While the model can identify most of the temporal regions involving the object, it is often unclear about which region is the correct moment and gives them similar confidence. This leads to low R1@0.3 and R1@0.5, whereas R5@0.3, R5@0.5 are often much higher. The above mentioned ambiguity can potentially be mitigated by clarifying text queries, such as specifying, ``Where was object X before I used it for the first time?''. 
%or by allowing a single text query to correspond to multiple correct moments.

%This ambiguity can be partly solved by improving the clarity of text queries, e.g., "Where was object X before I used it \textit{for the first time}?", 

%or allowing one text query to match with multiple correct moments.

% \TODO{
% - computation on Ego4D-Goalstep.
% - more hyper-parameter of Table 6.
% - limitations.
%  -- limitation in data/annotations.

% - using pretraining data of RGNet (before rebuttal).
% }

\end{document}